\documentclass[twoside,twocolumn]{paper}

\usepackage[toc,page]{appendix}

\usepackage{blindtext} 
\usepackage{amsmath}
\usepackage{bm}
\usepackage{graphicx}
\usepackage[T1]{fontenc}
\usepackage{tablefootnote}

\usepackage{microtype} \usepackage[T1]{fontenc} 
\usepackage[english]{babel} 
\usepackage[hmarginratio=1:1,top=32mm,width=190mm,columnsep=20pt]{geometry} \usepackage[hang, small,labelfont=bf,up,textfont=it,up]{caption} \usepackage{booktabs} 

\usepackage{enumitem} \setlist[itemize]{noitemsep} 
\usepackage{abstract}   

\usepackage{color}
\definecolor{dgray}{rgb}{0.8,0.8,0.8}
\definecolor{lgray}{rgb}{0.95,0.95,0.95}
\definecolor{nilscolor}{rgb}{.9,0.4,0.0}
\definecolor{nilscolB}{rgb}{.1,0.7,0.1}
\definecolor{commentcolor}{rgb}{0.2,0.3,0.8}

\usepackage{titling} \usepackage{hyperref}

\newcommand{\myrefeq}[1]{Eq.~(\ref{#1})}
\newcommand{\myreffig}[1]{Fig.~\ref{#1}}
\newcommand{\myreftab}[1]{Table~\ref{#1}}
\newcommand{\myrefsec}[1]{Section~\ref{#1}}
\newcommand{\myrefapp}[1]{Appendix~\ref{#1}}

\renewcommand{\v}[1]{{\bf{#1}}}
\newcommand{\vti}[1]{{\tilde{\bf{#1}}}}

\newcommand{\revi}[1]{#1}	
\newcommand{\reviB}[1]{#1}

\pretitle{\begin{center}\LARGE} 
\posttitle{\end{center}} 
\title{Deep Learning Methods for Reynolds-Averaged Navier-Stokes Simulations of Airfoil Flows}

\author{
\textsc{N. Thuerey, K. Wei\ss{}enow, L. Prantl, Xiangyu Hu} \\[0.8ex] 
\normalsize Technical University of Munich \\[1ex]
}

\date{} 

\begin{document}

\maketitle

\begin{abstract}
With this study, we investigate the accuracy of deep learning models 
for the inference of Reynolds-Averaged Navier-Stokes solutions.
We focus on a modernized U-net architecture and evaluate a large number of 
trained neural networks with respect to their accuracy for the calculation
of pressure and velocity distributions. In particular, we illustrate how training data 
size and the number of weights influence the accuracy of the solutions. 
With our best models, we arrive at a mean relative pressure and velocity error 
of less than 3\% across a range of previously unseen airfoil shapes. 
In addition all source code is publicly 
available in order to ensure reproducibility and to provide a starting point 
for researchers interested in deep learning methods for physics problems.
While this work focuses on RANS solutions, the neural network architecture 
and learning setup are very generic and applicable to a wide 
range of PDE boundary value problems on Cartesian grids.
\end{abstract}

\section{Introduction}

Despite the enormous success of deep learning methods in the field of 
computer vision \cite{krizhevsky2012imagenet,Isola2016,karras2017growgan},
and first success stories of applications in the area of 
physics simulations \cite{tompson2016accelerating,Xie2018,beck2018dlturb,raissi2018hidden,bar2018data},
the corresponding research communities retain a skeptical stance towards deep learning algorithms~\cite{durbin2018some}.
This skepticism is often driven by concerns 
about the accuracy achievable with deep learning approaches.
The advances of practical deep learning
algorithms have significantly outpaced the underlying theory \cite{yun2018critical}, 
and hence many researchers see these methods as black-box methods
that cannot be understood or analyzed. 

With the following study our goal is to investigate the accuracy
of trained deep learning models for the inference of Reynolds-averaged Navier-Stokes (RANS) 
simulations of airfoils in two dimensions.
We also illustrate that despite the lack of proofs, 
deep learning methods can be analyzed and employed thanks to the large number
of existing practical examples. We show how the accuracy of flow predictions
around airfoil shapes changes
with respect to the central training parameters, namely network size,
and the number of training data samples.
Additionally, we will demonstrate that the trained models yield a very high 
computational performance ''out-of-the-box''.

A second closely connected goal of our work is to provide a public testbed and evaluation platform
for deep learning methods in the context of computational fluid dynamics (CFD). 
Both code and training data are publicly available
at \url{https://github.com/thunil/Deep-Flow-Prediction}~\cite{webAirfoils},
and are kept as simple as possible to allow for quick adoption for experiments and further studies.
%
As learning task we focus on the direct inference of 
RANS solutions from a given choice of boundary conditions, i.e., airfoil shape and 
freestream velocity.
The specification of the boundary conditions as well as the solution of the flow problems 
will be represented by Eulerian field functions,
i.e. Cartesian grids. For the solution we typically consider
velocity and pressure distributions.
%
Deep learning as a tool makes sense in this setting, as 
the functions we are interested in, i.e. velocity and pressure,
are smooth and well-represented on Cartesian grids. Also, convolutional layers, as a particularly powerful 
component of current deep learning methods, are especially well suited for such grids.

The learning task for our goal is very simple when seen on a high level:
given enough training data, we have a unique relationship between boundary conditions and solution,
we have full control of the data generation process, very little noise in the solutions,
and we can train our models in a fully supervised manner. The difficulties rather stem 
from the non-linearities of the solutions, and the high requirements for accuracy.
To illustrate the inherent capabilities
of deep learning in the context of flow simulations we will also intentionally refrain from 
including any specialized physical priors such as conservation laws. Instead, we will employ 
straightforward, state-of-the-art convolutional neural network (CNN) architectures and evaluate 
in detail, based on more than 500 trained CNN models, 
how well they can capture the non-linear behavior of 
the Reynolds-averaged Navier-Stokes (RANS) equations.
As a consequence, the setup we describe in the following is a very generic approach for PDE boundary 
value problems, and as such is applicable to a variety of other equations beyond RANS.

\begin{figure*}[t!]
	\begin{center}
		\includegraphics[width=.9\textwidth]{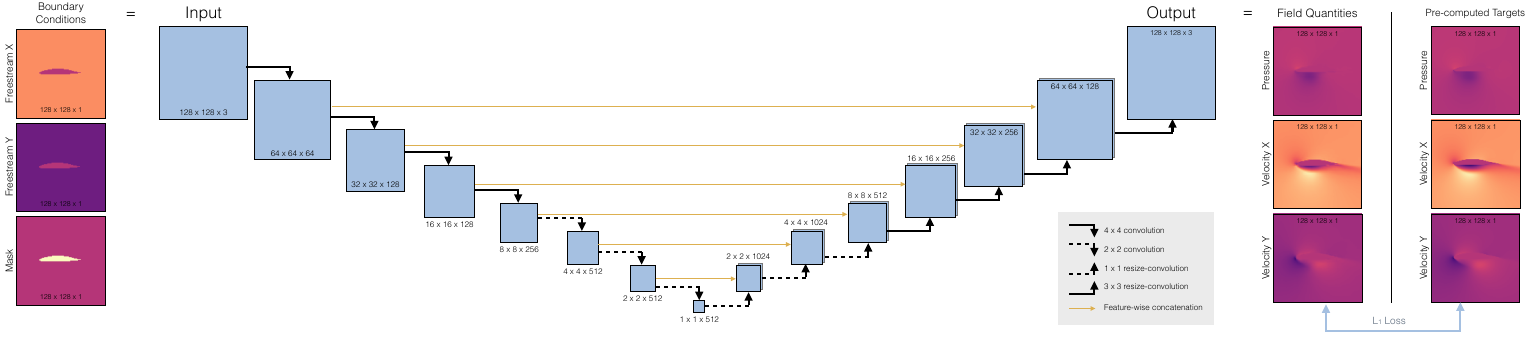}
	\end{center}
	\caption{ Our U-net architecture receives three constant fields as input, each containing the airfoil shape. 
	The black arrows denote convolutional layers, the details of which are given in \myrefapp{app:net},
	while orange arrows indicate skip connections.
	The inferred outputs have exactly the same size as the inputs, and are compared to the
	targets with an L$_1$ loss. 
	The target data sets on the right are pre-computed with OpenFOAM.}
	\label{fig:overview}
\end{figure*}

\section{Related Work}

Though machine learning in general has a long history in the field of CFD~\cite{greenman1999high},
more recent progresses mainly originate from the advent of deep learning, which can be
attributed to the seminal work of Krizhevsky et al.~\cite{krizhevsky2012imagenet}.
They were the first to employ deep CNNs in conjunction with GPU-based backpropagation training.
Beyond the original goals of computer vision \cite{schulman2013tracking,zheng2014detecting,jia20153d},
targeting physics problems with deep learning algorithms
has become a field of research that receives strongly growing interest.
Especially problems that involve dynamical systems pose highly interesting challenges.
Among them, several papers have targeted predictions of discrete Lagrangian systems.
E.g., Battaglia et al. \cite{battaglia2016interaction} introduced a network architecture to predict
two-dimensional rigid-body dynamics, that also can be employed for predicting object motions in videos \cite{watters2017visual}.
The prediction of rigid-body dynamics with a different architecture was proposed by Chang et al. \cite{chang2016compositional},
while improved predictions for Lagrangian systems were targeted by Yu et al.~\cite{yu2017long}.
Other researchers have used recurrent forms of neural networks (NNs) to predict 
Lagrangian trajectories for objects in height-fields \cite{ehrhardt2017learning}.

Deep learning algorithms have also been used to target a variety of flow problems,
which are characterized by continuous dynamics in high-dimensional Eulerian fields.
Several of the methods were proposed in numerical simulation, to speed up the solving process.
E.g., CNN-based pressure projections were proposed~\cite{tompson2016accelerating, yang2016data},
while others have considered learning time integration~\cite{wiewel2018}.
The work of Chu et al.~\cite{chu2017cnnpatch} targets an approach for increasing the resolution
of a fluid simulation with the help of learned descriptors. Xie et al.~\cite{Xie2018}
on the other hand developed a physically-based Generative Adversarial Networks 
(GANs) model for super-resolution flow simulations.
Deep learning algorithms also have potential for coarse-grained closure models~\cite{um2017mlflip,beck2018dlturb}.
Additional works have targeted the probabilistic learning of Scramjet predictions~\cite{soize2018enhancing},
and a Bayesian calibration of turbulence models for compressible jet flows~\cite{ray2018robust}.
Learned aerodynamic design models have also received attention for inverse problems
such as shape optimization~\cite{baque2018geodesic,umetani2018,sekar2019inverse}.

In the context of RANS simulations, deep learning techniques were successfully used
to address turbulence uncertainty~\cite{tracey2013application,edeling2014bayesian,ling2015evaluation},
as well as parameters for more accurate models~\cite{tracey2015machine}.
Ling et al., on the other hand, proposed a NN-based method to learn anisotropy tensors for 
RANS-based modeling~\cite{ling2016reynolds}.
Neural networks were also successfully used to improve turbulence models for
simulations of flows around airfoils~\cite{singh2017machine}.
While the aforementioned works typically embed a trained model in a numerical solver,
our learning approach directly infers solutions via the chosen CNN architecture.

The data-driven paradigm of deep learning methods has also led to algorithms
that learn reduced representations of space-time fluid data sets \cite{rtliquids2017,bkim2018deep}.
NNs in conjunction with proper orthogonal decompositions to obtain reduced representations 
of the flow were also explored~\cite{yu2018flowfield}.
Similar to our work, these approaches have the potential to yield new solutions very efficiently
by focusing on a known, constrained region of flow behavior. However, these  
works target the learning of reduced representations of the solutions, 
while we target the direct inference of solutions given a set of boundary 
conditions.

Closer to the goals of our work,
Farimani et al. have proposed a conditional GAN to infer numerical solutions  
of heat diffusion and lid-driven cavity problems~\cite{farimani2017}.
Methods to learn numerical discretization~\cite{bar2018data},
and to infer flow fields with semi-supervised learning~\cite{raissi2018hidden} were also proposed recently.
Zhang et al. developed a CNN, which infers the lift coefficient of 
airfoils~\cite{Zhang2017}. We target the same setting, but our networks aim for the calculation 
of high-dimensional velocity and pressure fields. 

While our work focuses on the U-Net architecture \cite{Ronneberger2015}, as shown in \myreffig{fig:overview},
a variety of alternatives has been proposed over the last years \cite{segnet,linknet,densenet}.
Adversarial training in the form of GANs~\cite{goodfellow2014generative,mirza2014conditional} is likewise very popular.
These GANs encompass a large class of modern deep learning methods 
for image super-resolution methods~\cite{ledig2016photo}, 
and for complex image translation problems~\cite{isola2016image,zhu2017cycle}.
Adversarial training approaches have also led to methods for the realistic synthesis of porous 
media \cite{mosser2017porous}, or point-based geometries~\cite{achlioptas2017representation}.
While GANs are a powerful concept, they are not beneficial in our setting, and 
we will focus on fully supervised training runs without discriminator networks.

In the following, we target solutions of RANS simulations.
Here, a variety of application areas \cite{knight1998shock,shur1999detached,walters2002new,bueno2012continuous}, 
hybrid methods \cite{frohlich2008hybrid}, and
modern variants~\cite{rung2003restatement,gerolymos2004contribution,poroseva2014velocity}
exists. We target the classic Spalart-Allmaras (SA) model \cite{spalart1992one}, as 
this type of solver represents a well-established
and studied test-case with practical industry relevance.

\section{Non-linear Regression with Neural Networks}

Neural networks can be seen as a general methodology to regress arbitrary
non-linear functions $f$. In the following, we give a very brief
overview. More in depth explanations can be found in 
corresponding books \cite{bishop2006book,Goodfellow2016}.

We consider problems of the form $\v{y} = \hat{f}( \v{x} )$,
i.e., for given an input $\v{x}$ we want to approximate the output $\v{y}$ of
the true function $\hat{f}$ as closely as possible
with a representation $f$ based on the degrees of freedom $\v{w}$ such that $\v{y} \approx f( \v{x} , \v{w} )$.
In the following, we choose neural networks (NNs) to represent $f$.
Neural networks model the target functions with networks of nodes that are connected to one another.
Nodes typically accumulate values from previous nodes, and apply
so-called activation functions $g$. These activation functions are crucial to introduce non-linearity,
and effectively allow NNs to approximate arbitrary functions. 
Typical choices for $g$ are hyperbolic tangent, sigmoid and ReLU functions.

\newcommand{\nlll}{l}
The previous description can be formalized as follows: for a layer $\nlll$ in the network, 
the output of the i'th node $a_{i,\nlll}$ 
is computed with 
\begin{equation} \label{eq:nnLayer}
	a_{i,\nlll} = g \Big( \sum_{j=0}^{n_{\nlll-1}} w_{ij,\nlll-1} \ a_{j,\nlll-1} \Big) \ .
\end{equation}
Here, $n_\nlll = $ denotes number of nodes per layer. 
To model the bias, i.e., a per node offset, we assume $a_{0,\nlll} = 1$ for all $\nlll$.
This bias is crucial to allow nodes to shift the input to the activation function. 
We employ this commonly used formulation to denote all degrees of
freedom with the weight vector $\v{w}$. 
Thus, we will not explicitly distinguish regular weights and biases below.
We can rewrite \myrefeq{eq:nnLayer} using a weight matrix $W$ as
$\mathbf{a}_{\nlll} = \mathbf{g} ( W_{\nlll-1} \mathbf{a}_{\nlll-1}) $.
In this equation, $\mathbf{g}$ is applied component-wise to the input vector.
Note that without the non-linear activation functions $\mathbf{g}$ we could represent a whole 
network with a single matrix $W_0$, i.e., $\mathbf{a} = W_0 \mathbf{x}$. 

To compute the weights, we have to provide 
the learning process with a loss function  $L( \v{y} , f( \v{x} , \v{w} ) )$.
This loss function is problem specific, and typically has the dual goal to evaluate the 
quality of the generated output with respect to $\v{y}$, as well as 
reduce the potentially large space of solutions by regularization. 
The loss function $L$ needs to be at least once differentiable,
so that its gradient $\nabla_{y}L$ can be back-propagated into the network
in order to compute the weight gradient $\nabla_{w}L$.

Moving beyond fully connected layers, where all nodes of two adjacent layers are densely connected,
so called {\em convolutional layers} are central components that drive many 
successful deep learning algorithms.
These layers make use of fixed spatial arrangements of the input data, in order to
learn filter kernels of convolutions. 
Thus, they represent a subset of fully connected layers typically with much fewer weights. 
It has been shown that these convolutional layers
are able to extract important {\em features} of the input data, and each convolutional layer typically
learns a whole set of convolutional kernels.

The convolutional kernels typically have only a small set of weights, e.g., 
$n\!\times\!n$ with $n=5$ for a two-dimensional data set. 
As the inputs typically consist of vector quantities, e.g., $i$ different channels of data,
the number of weights for a convolutional layer with $j$ output features is $n^2\!\times\!i\!\times\!j$,
with $n$ being the size of the kernel. These convolutions extend naturally to higher dimensions.

In order to learn and extract features with larger spatial extent, it is typically preferable to 
reduce or enlarge the size of the inputs rather than enlarging the kernel size. For these resizing
operations, NNs commonly employ either {\em pooling} layers or strided convolutions. While strided convolutions
have the benefit of improved gradient propagation, pooling can have advantages for smoothness 
when increasing the spatial resolution (i.e. ''de-pooling'' operations) \cite{odena2016deconv}.
Stacks of convolutions in conjunction with changes of the spatial size of the input data
by factors of two, are common building blocks of many modern NN architectures. 
Such convolutions acting on different spatial scales have significant benefits over fully
connected layers, as they can lead to vastly reduced weight numbers and correspondingly
well regularized convolutional kernels. Thus, the resulting smaller networks are easier to train,
and usually also have reduced requirements for the amounts of training data that is needed
to reach convergence. 

\section{Method}
\label{sec:method}

In the following, we describe our methodology for the deep learning-based inference
of RANS solutions. 

\subsection*{Data Generation}

\begin{figure*}[bt]
	\begin{center}
		\includegraphics[width=0.95\textwidth]{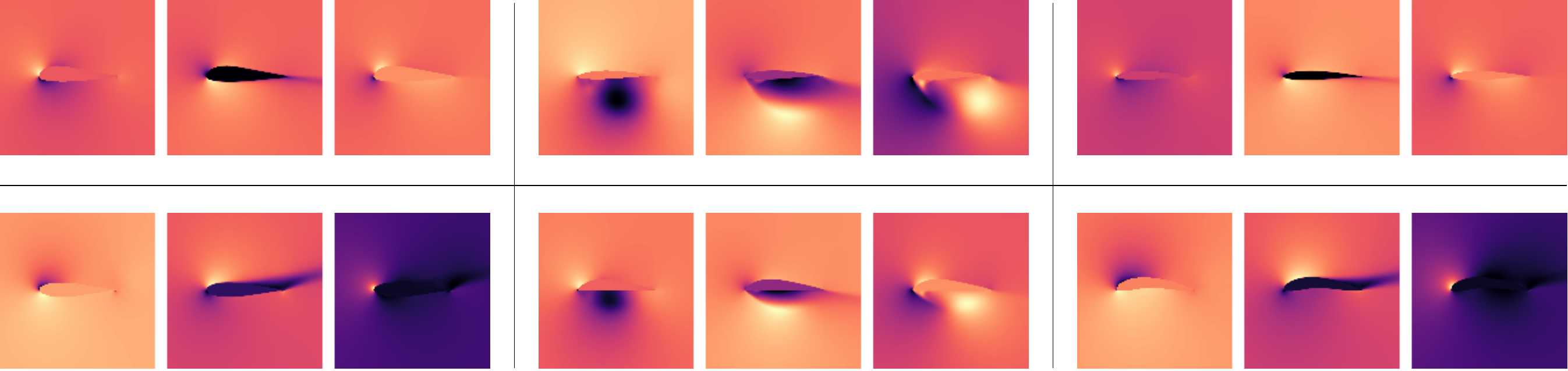}
	\end{center}
	\caption{A selection of simulation solutions used as learning targets. Each triple
	contains pressure, x, and y components of the velocity. Each component and data set is normalized independently
	for visualization, thus these images show the structure of the solutions
	rather than the range of values across different solutions. }
	\label{fig:refExamples}
\end{figure*}

In order to generate ground truth data for training,
we compute the velocity and pressure distributions of flows around airfoils.
We consider a space of solutions with a 
range of Reynolds numbers $Re = [0.5 , 5]$ million, incompressible flow,
and angles of attack in the range of $\pm 22.5$ degrees.
We obtained 1505 different airfoil shapes from the UIUC database \cite{uiuc1996}, 
which were used to generate input data in conjunction with randomly sampled 
freestream conditions from the range described above. 
The RANS simulations make use of the widely used SA~\cite{spalart1992one}
one equation turbulence model, and solutions are calculated with 
the open source code {\em OpenFOAM}. 
%
Here we employ a body-fitted triangle mesh, with refinement near the airfoil.
At the airfoil surface, the triangle mesh has an average edge length of $1/200$
to resolve the boundary layer of the flow. The discretization is coarsened to an edge length of
$1/16$ near the domain boundary, which has an overall size of $8 \times 8$ units. 
The airfoil has a length of $1$ unit.
Typical examples from the training data set are shown in \myreffig{fig:refExamples}.

\reviB{
While typical RANS solvers, such as the one from
OpenFOAM, require large distances for the domain boundaries in order to reduce their 
negative impact on the solutions around the airfoil, 
we can focus on a smaller region of $2 \times 2$ units 
around the airfoil for the deep learning task. As our model directly learns from 
reference data sets that were generated with a large domain boundary distance, we do 
not need to reproduce solution in the whole space where the solution was computed, but rather 
can focus on the region in the vicinity of the airfoil without impairing the solution. 
}

\begin{figure}[bt]
	\begin{center}
	\includegraphics[width=0.495\textwidth]{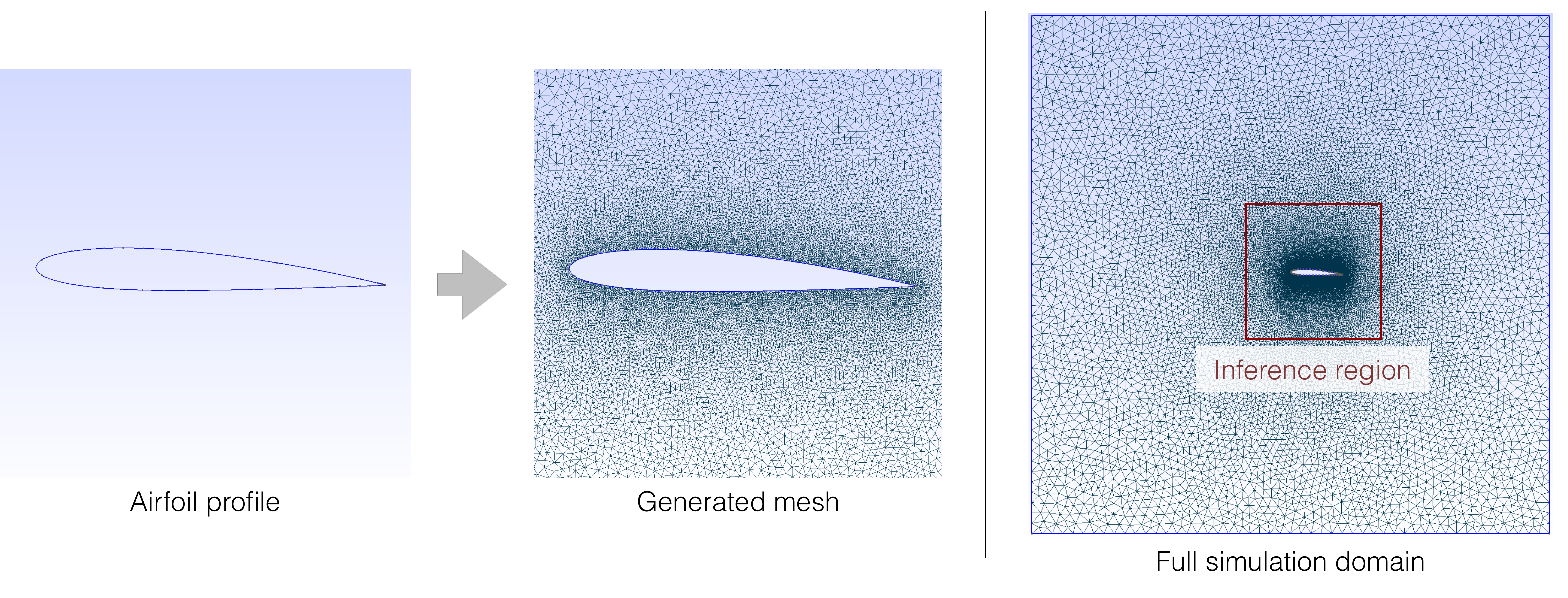}
	\end{center}
	\caption{An example UIUC database entry with the corresponding simulation mesh.
	The region around the airfoil considered for CNN inference is highlighted with a red square. }
	\label{fig:mesh}
\end{figure}

\revi{
To facilitate NN architectures with convolutional layers, we resample this region 
with a Cartesian $128^2$ grid to obtain the ground truth pressure and velocity data sets, as shown 
in \myreffig{fig:mesh}. The re-sampling is performed with 
a linear weighted interpolation of cell-centered values with a spacing 
of $1/64$ units in OpenFOAM.} \reviB{
As for the domain boundaries, we only need to ensure that the original solution was produced with a sufficient 
resolution to resolve the boundary layer. As the solution is smooth, we can later on sample it with 
a reduced resolution. Both properties, the reduced spatial extent of the deep learning region and the 
relaxed requirements for the discretization, highlight advantages of deep learning methods over 
traditional solvers in our setting. 
}

\revi{
We randomly choose an airfoil, Reynolds number and angle of attack from the parameter ranges 
described above and compute the corresponding RANS solution
to obtain data sets for the learning task. 
This set of samples is split into two parts: a larger fraction that is used for training,
and a remainder that is only used for evaluating the current state of a model (the {\em validation} set).
Details of the respective data set sizes are given in the appendix, we typically use an 80\% to 20\% split 
for training and validation sets, respectively. The validation set allows for 
an unbiased evaluation of the quality of the trained model during training, e.g., to detect overfitting.
}

To later on evaluate the capabilities of the trained models with respect to generalization,
we use an additional set of 30 airfoil shapes that were not used for training,
to generate a {\em test data set} with 90 samples (using the
same range of Reynolds numbers and angles of attack as described above).

\subsection*{Pre-processing}
\label{sec:prepro}

As the resulting solutions of the RANS simulations have a size of $128^2 \times 3$,
we use CNN architectures with inputs of the same size. The solutions globally
depend on all boundary conditions. Accordingly, the architecture of the network 
makes sure this information is readily available spatially and throughout the different layers.

Thus freestream conditions 
and the airfoil shape are encoded in a $128^2 \times 3$ grid of values.
As knowledge about the targeted Reynolds number is required to compute the desired
output, we encode the Reynolds number in terms of differently scaled freestream
velocity vectors. I.e., for the smallest Reynolds numbers the freestream velocity has a magnitude 
of 0.1, while the largest ones are indicated by a magnitude of 1.
The first of the three input channels contains a 
$[0,1]$ mask $\phi$ for the airfoil shape, 0 being outside, and 1 inside. The next two channels
of the input contain x and y velocity components, $\v{v}_{i} = (v_{i,x},v_{i,y})$ respectively. 
Both velocity channels are initialized to the x and y component of the freestream conditions, respectively,
with a zero velocity inside of the airfoil shape.
Note that the inputs contain highly redundant information, they are essentially constant, 
and we likewise, redundantly, encode the airfoil shape in all three input fields.

The output data sets for supervised training have the same 
size of $128^2 \times 3$. Here, the first channel contains pressure $p$, while
the next two channels contain x and y velocity of the RANS solution, $\v{v}_{o} = (v_{o,x},v_{o,y})$.

While the simulation data could be used for training in this form, we describe two further data 
pre-processing steps that we will evaluate in terms of their influence on the learned performance 
below. 
First, we can follow common practice and normalize all involved quantities with respect to the magnitude of the freestream 
velocity, i.e., make them dimensionless. 
Thus we consider $\vti{v}_{o} = \v{v}_{o}/|\v{v}_{i}|$, and $\tilde{p}_{o} = p_{o}/|\v{v}_{i}|^2$.
Especially the latter is important to remove the quadratic scaling of the pressure values from the 
target data. This flattens the space of solutions, and simplifies the task for the neural network later on.

In addition, only the pressure gradient is typically needed to compute the RANS solutions.
Thus, as a second pre-processing variant we can additionally remove the
mean pressure from each solution and define $\hat{p}_{o} = \tilde{p}_{o} - p_{\text{mean}}$,
with the pressure mean $p_{\text{mean}} = \sum_i p_i / n$, where $n$ denotes the number of 
individual pressure samples $p_i$. 
Without this removal of the mean values, the pressure targets represent an ill-posed learning goal
as the random pressure offsets in the solutions are not correlated with the inputs.

As a last step, irrespective of which pre-processing method was used,
each channel is normalized to the $[-1,1]$ range in order
to minimize errors from limited numerical precision during the training phase. 
We use the maximum absolute value for 
each quantity computed over the entire training data set to normalize the data. Both boundary 
condition inputs and ground truth solutions, i.e. output targets,
are normalized in the same way. Note that the velocity is not offset, but only scaled by 
a global per component factor, even when using $\hat{p}_{o}$.

\subsection*{Neural Network Architecture}

Our neural network model is based on the U-Net architecture \cite{Ronneberger2015}, 
which is widely used for tasks such as image translation. The network has the typical bowtie 
structure, translating spatial information into extracted features
with convolutional layers. In addition,  
skip-connections from in- to output feature channels are introduced to 
ensure this information is available in the outputs layers for inferring the solution.
We have experimented with a variety of architectures with different amounts 
of skip connections \cite{segnet,linknet,densenet}, and found that the U-net yields a very good quality
with relatively low memory requirements.
Hence, we will focus on the most successful variant, a modified U-net architecture in the following.

This U-Net is a special case of an encoder-decoder architecture.
In the encoding part, the image size is progressively down-sampled by a factor of 2 
with strided convolutions. The allows the network to extract increasingly large-scale
and abstract information in the growing number of feature channels. The decoding part of
the network mirrors this behavior, and increases the spatial resolution with average-depooling
layers, and reduces the number of feature layers. The skip connections
concatenate all channels from the encoding branch to the corresponding branch of the decoding part,
effectively doubling the number of channels for each decoding block. These skip
connections help the network to consider low-level input information during the 
reconstruction of the solution in the decoding layers.
Each section of the network consists of a convolutional
layer, a batch normalization layer, in addition to a non-linear activation function. 
For our standard U-net with 7.7m weights, we use 7 convolutional blocks to turn the $128^2\times 3$ 
input into 
a single data point with 512 features, typically using convolutional kernels of size $4^2$
(only the inner three layers of the encoder use $2^2$ kernels, see \myrefapp{app:net}). 
As activation functions we use leaky ReLU functions with a slope of 0.2 in the encoding layers, and 
regular ReLU activations in the decoding layers.
The decoder part uses another 7 symmetric layers to reconstruct the target 
function with the desired dimensionality of $128^2\times 3$.

While it seems wasteful to repeat the freestream conditions almost $128^2$ times, i.e., over the 
whole domain outside of the airfoil, 
this setup is very beneficial for the NN. We know that the 
solution everywhere depends on the boundary conditions, and while the network would eventually 
learn this and propagate the necessary information via the convolutional bowtie structure, it
is beneficial for the training process to make sure this information is available everywhere right 
from the start. This motivates the architecture with redundant boundary condition information and 
skip connections.

\begin{figure*}[tb]
	\begin{center}
		\includegraphics[width=0.9 \textwidth]{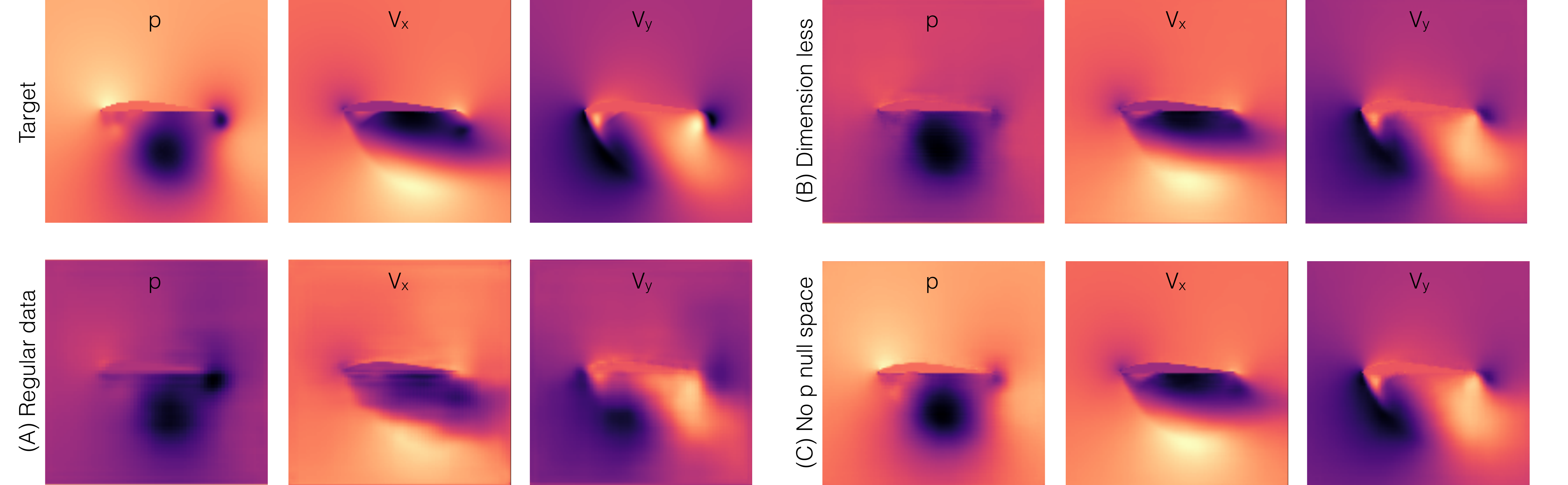}
	\end{center}
	\caption{Ground truth target, and three different data pre-processing variants.
	It is clear that using the data directly (A) leads to strong artifacts
	with blurred and jagged solutions. While velocity 
	normalization (B) yields significantly better results, the pressure values still show 
	large deviations from the target. These are reduced by removing the pressure null space from 
	the data (C).}
	\label{fig:prepro}
\end{figure*}

\subsection*{Supervised Training}

We train our architecture with the \emph{Adam} optimizer \cite{kingma2014adam},
using 80k iterations unless otherwise noted. 
Our implementation is based on the {\em PyTorch}\footnote{From {\em https://pytorch.org}.} deep learning framework.
Due to the strictly supervised setting of our learning setup,
we use a simple L$_1$ loss $L = |\v{y}-\v{a}|$, 
with $\v{a}$ being the output of the CNN.
\revi{Here, an L$_2$ loss could likewise be used and yields very similar results, 
but we found that L$_1$ yields slight improvements. 
In both cases, a supervision for all cells of the inferred outputs 
in terms of a direct vector norm is important for stable training runs.
The number of iterations was chosen such that the training runs converge 
to their final inference accuracies across all changes of hyperparameters and network
architectures. Hence, to ensure that the different training runs below can be compared, 
we train all networks with the same number of training iterations.}

\begin{figure}[bt]
	\begin{center}
		\includegraphics[width=0.23\textwidth]{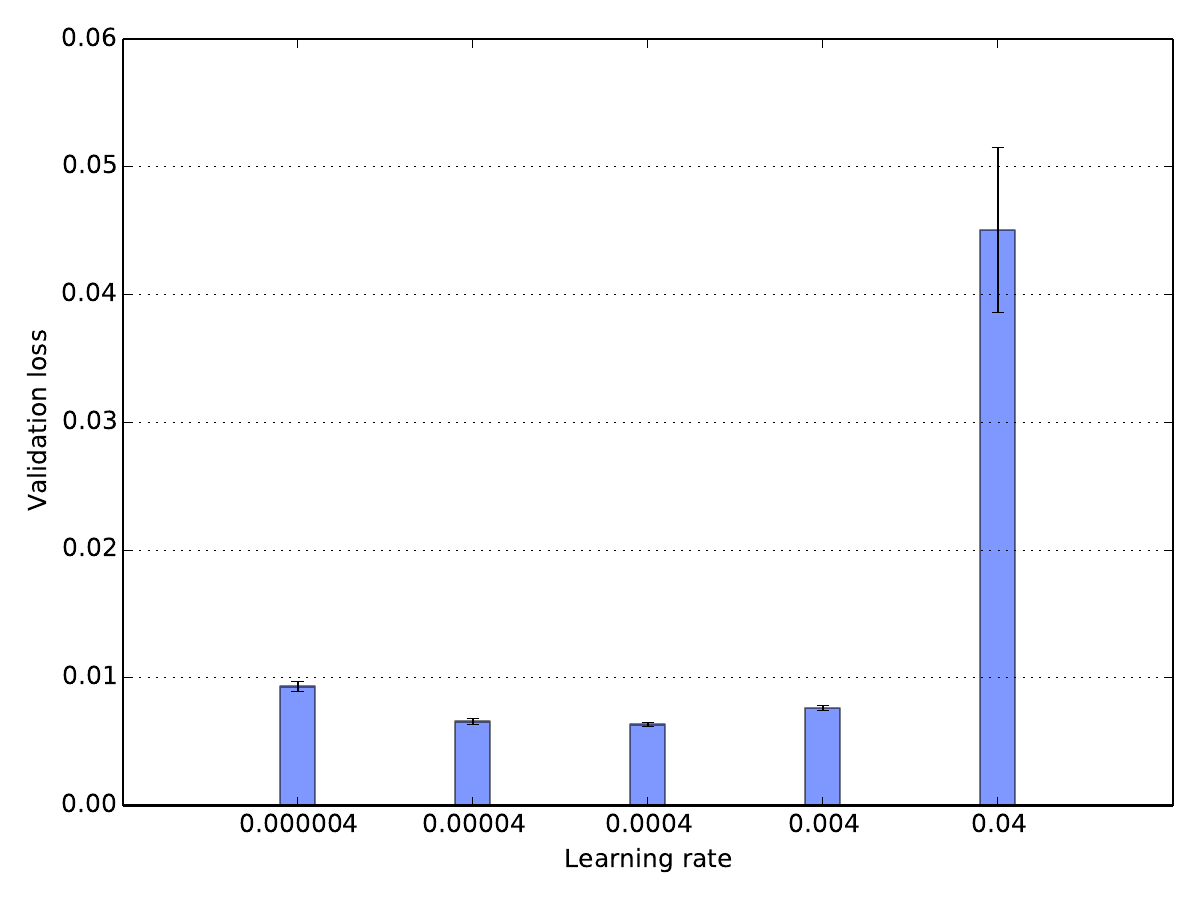} \ 
		\includegraphics[width=0.23\textwidth]{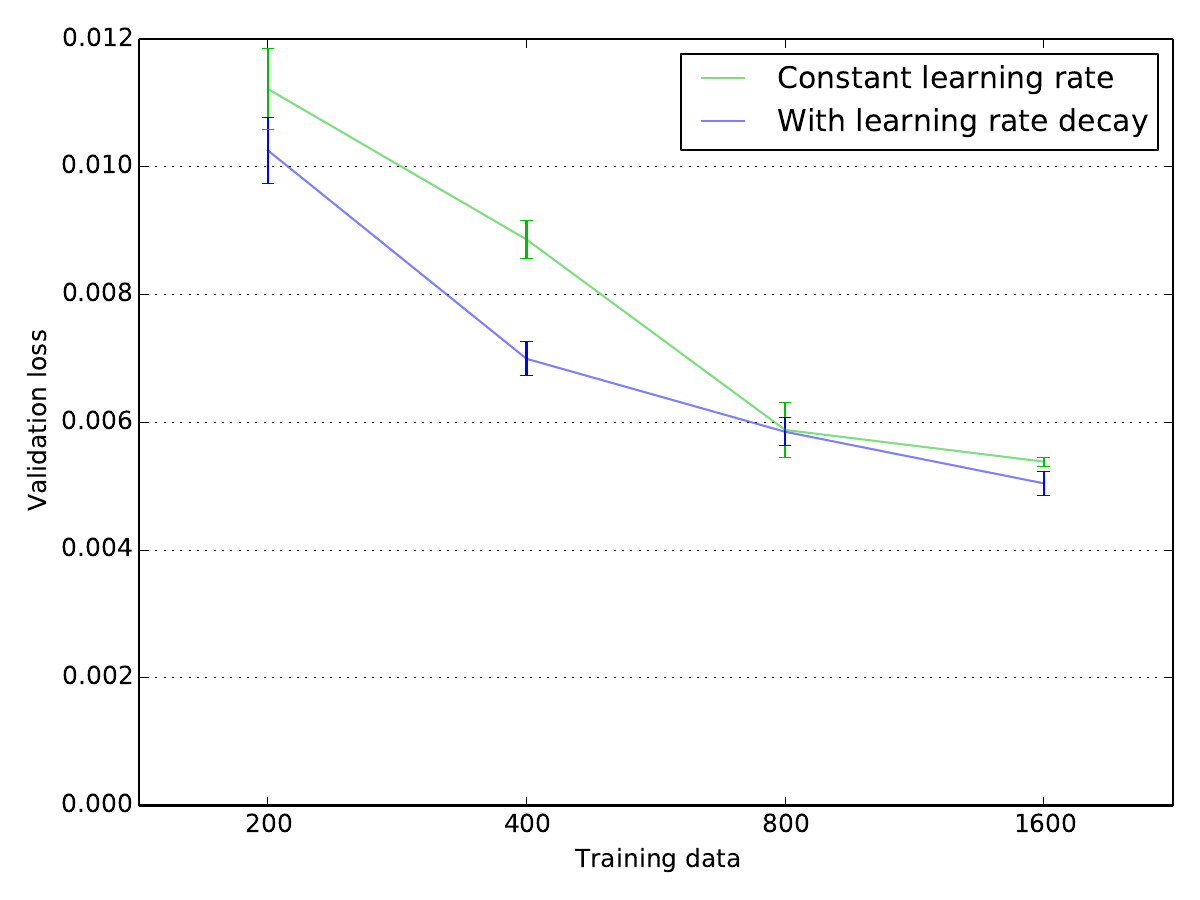}
		\caption{
		Left: Varied learning rates shown in terms of validation loss for 
		an otherwise constant learning problem.
		Right: Validation loss for a series of models with and without
		learning rate decay (amount of training data is varied). 
		The decay helps to reduce variance and performance
		especially when less training data is used.
		}
		\label{fig:learningRates}
	\end{center}
\end{figure}

\begin{figure}[bt]
	\begin{center}
		\includegraphics[width=0.4\textwidth]{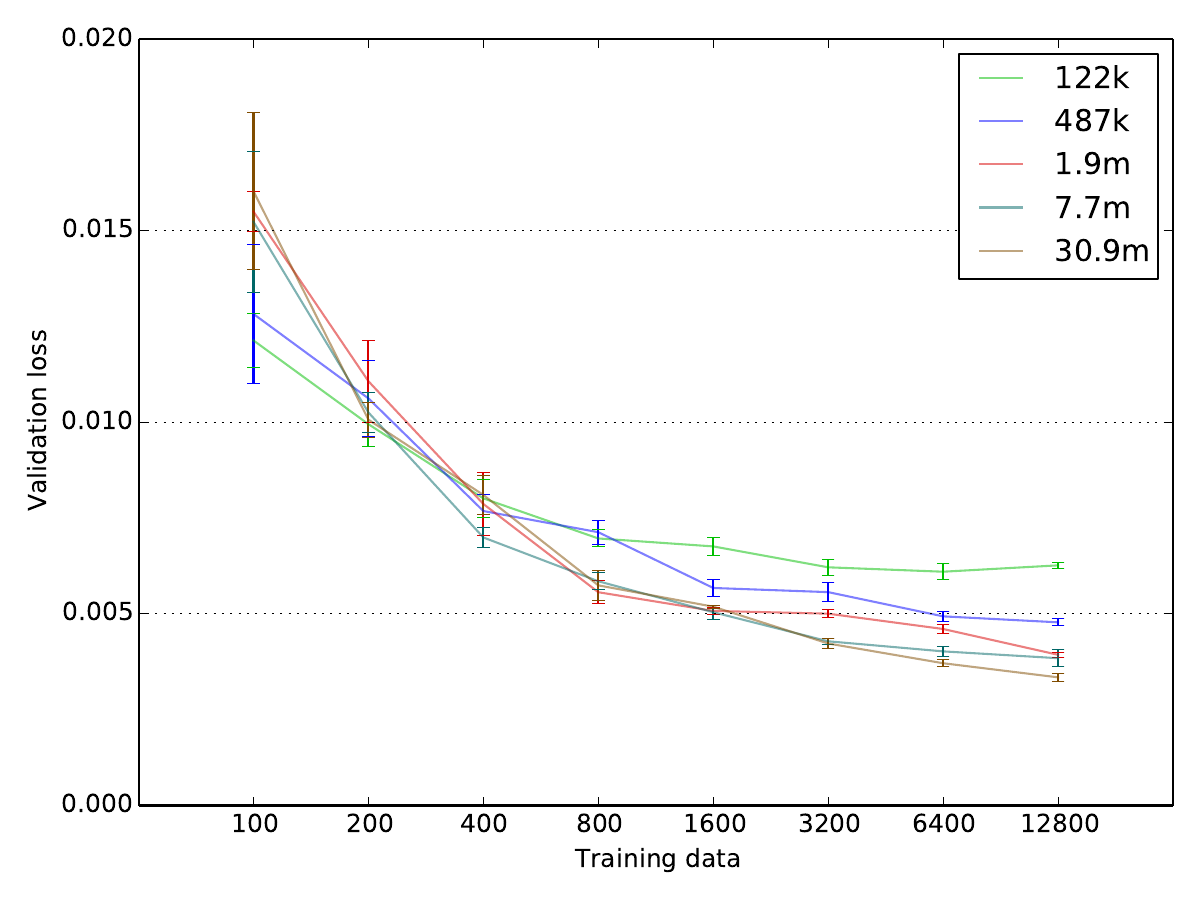}
		\includegraphics[width=0.4\textwidth]{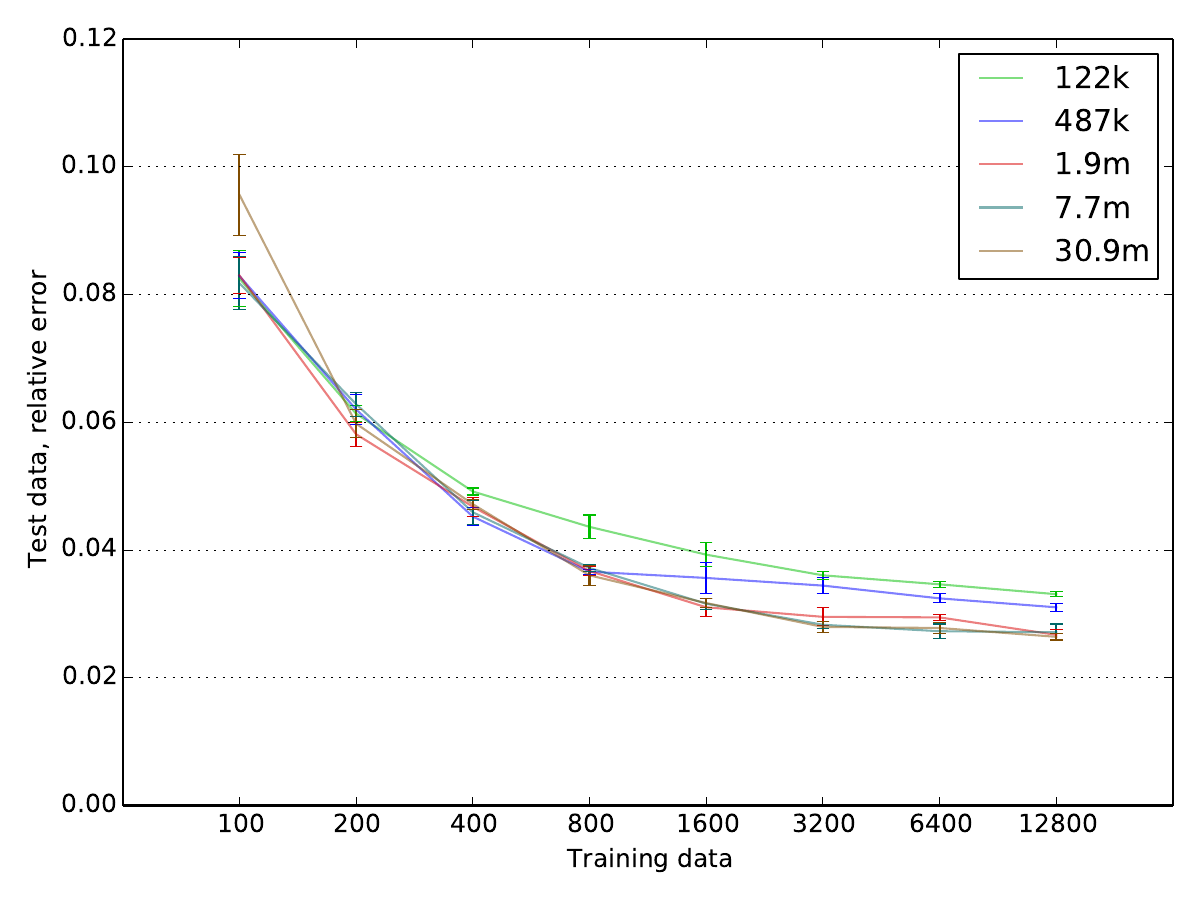}
		\caption{Validation and testing accuracy for different model sizes and
		training data amounts.}
		\label{fig:expcomp}
	\end{center}
\end{figure}

\revi{Due to the potentially large number of local minima and the stochasticity 
of the training process, individual runs can
yield significantly different results due to effects such as non-deterministic
GPU calculations and/or different random seeds.}
While runs can be made deterministic,
slight changes of the training data or their order can lead to similar 
differences in the trained models. 
\revi{
Thus, in the following we present network performances
across multiple training runs. For practical applications, a single, best performing 
model could be selected from such a collection of runs via cross validation.}
The following graphs, e.g. \myreffig{fig:learningRates} and 
\myreffig{fig:expcomp}, show the mean and standard error of the mean for five runs, all with otherwise 
identical settings apart from different random seeds. Hence, the standard errors indicate the 
variance in result quality that can be expected for a selected training modality.

First, we illustrate the importance of proper data normalization. As outlined above 
we can train models either (A) with the pressure and velocity data exactly as they arise 
in the model equations, i.e., $(\v{v}_o,p_o)$, or (B) we can normalize the data by 
velocity magnitude $(\vti{v}_o,\tilde{p}_o)$. Lastly, we can remove the pressure null space 
and train models with $(\vti{v}_o,\hat{p}_o)$ as target data (C). Not surprisingly,
this makes a huge difference. For comparing the different variants, we 
de-normalize the data for (B) and (C), and then compute the averaged, absolute error 
w.r.t. ground truth pressure and velocity values for 400 randomly selected data sets.
While variant (A) exhibits a very significant average error of $291.34$, the 
data variant (B) has an average error of only $0.0566$, while (C) reduces this by 
another factor of ca. 4 to $0.0136$.
An airfoil configuration that shows an example of how these errors manifest themselves
in the inferred solutions can be found in \myreffig{fig:prepro}.

Thus, in practice it is crucial to understand the data that the model should learn,
and simplify the space of solutions as much as possible, e.g., by 
making use of the dimensionless quantities for fluid flow.
Note that for the three models discussed above we have already used 
the training setup that we will explain in more detail in the following paragraphs. From now on, 
all models will only be trained with fully normalized data, i.e., case (C) above.

We have also experimented with adversarial training, and while others
have noticed improvements~\cite{farimani2017}, we were not successful with GANs during our
tests. While individual runs yielded good results, this was usually caused by sub-optimal 
settings for the corresponding supervised training runs.
In a controlled setting such as ours, where we can densely sample the 
parameter space, we found that generating more training data,
rather than switching to a more costly adversarial training,
is typically a preferred way to improve the results.

\paragraph{Basic Parameters}
In order to establish a training methodology, we first evaluate several
basic training parameters, namely learning rate, learning rate decay, 
and normalization of the input data. 
In the following we will evaluate these parameters for 
a standard network size with a training data set of medium size 
(details given in \myrefapp{app:data}). 

One of the most crucial parameters for deep learning is the learning rate
of the optimizer. 
\revi{
The learning rate scales the step size the optimizer takes to update the weights of a neural network
based on a gradient computed from one mini-batch of data.
As the energy landscapes spanned by typical deep neural networks
are often non-linear functions with large numbers of minima and saddle-points,
small learning rates are not necessarily ideal, as the optimization might get stuck 
in undesirable states, while overly large ones can easily prevent 
convergence.} \myreffig{fig:learningRates} illustrates this for our setting. 
The largest learning rate clearly overshoots and has trouble converging. In our case 
the range of $4\cdot10^{-3}$ to $4\cdot10^{-4}$ yields good convergence.

In addition, we found that learning rate decay, i.e., decreasing the learning rate 
during training, helps to stabilize the results, and reduce variance in performance. 
While the influence is not huge when the other parameters are chosen well, we decrease 
the learning rate to 10\% of its initial value over the course of the second half 
of the training iterations. 
A comparison of four different settings each with and without learning rate
decay is shown on the right of \myreffig{fig:learningRates}.

\revi{
To prevent overfitting, i.e., the reproduction of single input-output pairs rather 
than a smooth reproduction of the function that should be approximated by the model,
regularization is an important topic for deep learning methods. The most common regularization techniques 
are dropout, data augmentation, and early stopping \cite{Goodfellow2016}. In addition,
the loss function can be modified, typically with additional terms that minimize 
parts of the outputs or even the NN weights ({\em weight decay}) in terms of L$_1$ or L$_2$ norms.
While this regularization is especially important for fully connected 
neural networks \cite{ling2016reynolds,singh2017machine}, we focus on CNNs 
in our study, which are inherently less prone to overfitting as they are applied 
to all spatial locations of an input, and typically receive a large variety of data configurations
even from a single input to the NN.} 

In addition to the techniques outlined above, 
overfitting can also be avoided by ensuring that enough training data 	
is available. While this is not always possible or practical, 
a key advantage of machine learning in the context of PDEs is that reliable training data 
can be generated in large amounts given enough computational resources.
We will investigate the influence of varying amounts of training data for our models in more 
detail below. While we found a very slight amount of dropout to be preferable 
(see \myrefsec{app:net}), we will not use any other regularization methods.
We found that our networks converged to stable 
levels in terms of training and validation losses, and hence do not employ
early stopping in order to ensure all networks below were trained with the 
same number of iterations. It is also worth noting that data augmentation 
is difficult in our context, as each solution is unique for a given input configuration.
Below, we instead focus on the influence of varying amounts of pre-computed training data on 
the generalizing capabilities of a trained network. 

Note that due to the complexity of typical learning tasks for NNs it
is non-trivial to find the best parameters for training. 
Typically, it would be ideal 
to perform broad hyperparameter searches for each of the different 
hyperparameters to evaluate their effect on inference accuracy.
However, as differences of these searches were found to be relatively small in our case, 
the training parameters will be kept constant:
the following training runs use a learning rate of 
0.0004, a batch size of 10, and learning rate decay.

\paragraph{Accuracy}

Having established a stable training setup, we can now investigate the accuracy 
of our network in more detail. Based on a database of 26.722 target solutions
generated as described above,
we have measured how the amount of available training data influences accuracy 
for validation data as well as generalization accuracy for a test data set. 
In addition, we measure how accuracy scales with the number of weights, i.e., degrees of freedom in the NN.

For the following training runs, the amount of training data is scaled from 100 samples to 12800 samples in factors of two, 
and we vary the number of weights by scaling the number of feature maps in the convolutional layers of our network.
As the size of the kernel tensor of a convolutional layer scales with the number of input channels times 
number of output channels, a 2x increase of channels leads to a roughly four-fold increase in overall weights
(biases change linearly).
The corresponding accuracy graphs for five different network sizes can be seen in \myreffig{fig:expcomp}. 
As outlined above, five models with different random seeds (and correspondingly
different sets of training data) where trained for each of the data points, standard errors are shown with error 
bars in the graphs. The different networks have 122.979, 487.107, 1.938.819, 7.736.067,
and 30.905.859 weights, respectively.
The validation loss in  \myreffig{fig:expcomp},top shows how the models with little amounts of training data 
exhibit larger errors, and vary very significantly in terms of performance. The behavior stabilizes with larger 
amounts of data being available for training, and the models saturate in terms of inference accuracy at different 
levels that correspond to their weight numbers. Comparing the curves for the 122k and 30.9m models, the former
exhibits a flatter curve with lower errors in the beginning (due to inherent regularization from the smaller
number of weights), and larger errors at the end. The 30.9m instead more strongly overfits to the data initially,
and yields a better performance when enough data is available. In this case the mean error for the latter model
is 0.0033 compared to 0.0063 for the 122k model.

The graphs also show how the different models start to saturate in terms of loss reduction once a certain
amount of training data is available. For the smaller models this starts to show around 1000 samples, while 
the trends of the larger models indicate that they could benefit from even more training data.
The loss curves indicate that despite the 4x increase in weights for the different networks, 
roughly doubling the amount of training data is sufficient to reach a similar saturation point.

\begin{figure}[bt]
	\begin{center}
		\includegraphics[width=0.4\textwidth]{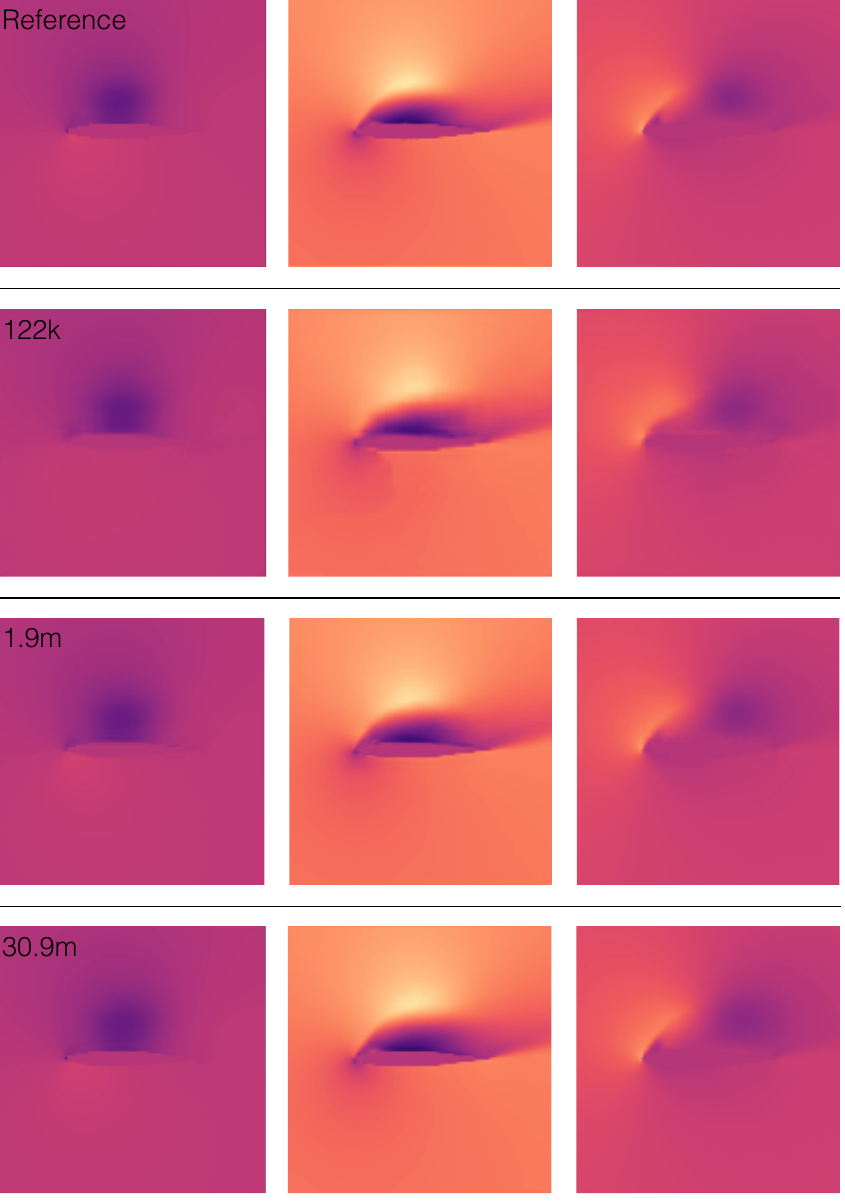}
		\caption{An example result from the test data set, with reference at the top, and 
		inferred solutions by three different model sizes (122k,1.9m, and 30.9m).
		The larger networks produce visibly sharper and more detailed solutions.}
		\label{fig:expcompPic}
	\end{center}
\end{figure}

The bottom graph of \myreffig{fig:expcomp} shows the performance of the different models for a test set 
of 30 airfoils that were not seen during training. 
Freestream velocities were randomly sampled from the same distribution used 
for training data generation (see \myrefsec{sec:method}) to produce 90 test data sets.
We use this data to measure how well the trained models 
can generalize to new airfoil shapes. Instead of the L$_1$ loss that was used for the validation data,
this graph shows the mean relative error of the normalized quantities 
(the L$_1$ loss behavior is line with the relative errors shown here).
The relative error is computed with $e_f = 1 / n \sum_i^n |\tilde{f} - f| / |f|$, with $n$ being the number of 
samples, $128^2$ in our case, $f$ the function under consideration, e.g., pressure, and 
$\tilde{f}$ the approximation of the neural network.
We found the average relative error for all inferred fields, i.e. $e_{avg} = (e_p + e_{v_{o,x}} + e_{v_{o,y}})/3$ 
to be a good metric to evaluate the models, as it takes all outputs into account and
directly yields an error percentage for the accuracy of the inferred solutions. In addition, it sheds light 
on the relative accuracy of the inferred solutions, while the L$_1$ metric yields  
estimates of the averaged differences. Hence, both metrics are important for evaluating the overall 
accuracy of the trained models.

For this test set, the curves exhibit a similar fall-off with reduced 
errors for larger amounts of training data, but the difference between the different model sizes is
significantly smaller. With the largest amount of training data (12.8k samples), the difference 
in test error between smallest and largest models is 0.033 versus 0.026. 
Thus, the latter network achieves an average relative error of 2.6\% across all three output channels.
Due to the differences between velocity and pressure functions, this error is not evenly distributed.
Rather, the model is trained for reducing $L_1$ differences across all three output quantities, 
which yields relative errors of 2.15\% for the x velocity channel, 2.6\% for y, and 14.76\% for pressure values.
Especially the relatively large amount of small pressure values with fewer
large spikes in the harmonic functions lead to increased relative errors for the pressure channel.
If necessary, this could be alleviated by changing the loss function, but as 
the goal of this study is to consider generic CNN performance, we will 
continue to use $L_1$ loss in the following.
 
Also, it is visible in \myreffig{fig:expcomp}
that the three largest model sizes yield a very similar performance. While the models improve in terms 
of capturing the space of training data, as visible from the validation loss at the top of \myreffig{fig:expcomp}, 
this does not directly translate into an improved generalization. 
The additional training data in this case does not yield new information for 
the unseen shapes. An example data set with inferred solutions is visualized 
in \myreffig{fig:expcompPic}. Note that the relatively small numeric changes of the
overall test error lead to significant differences in the solutions.

\begin{figure}[bt]
	\begin{center}
		\includegraphics[width=0.4\textwidth]{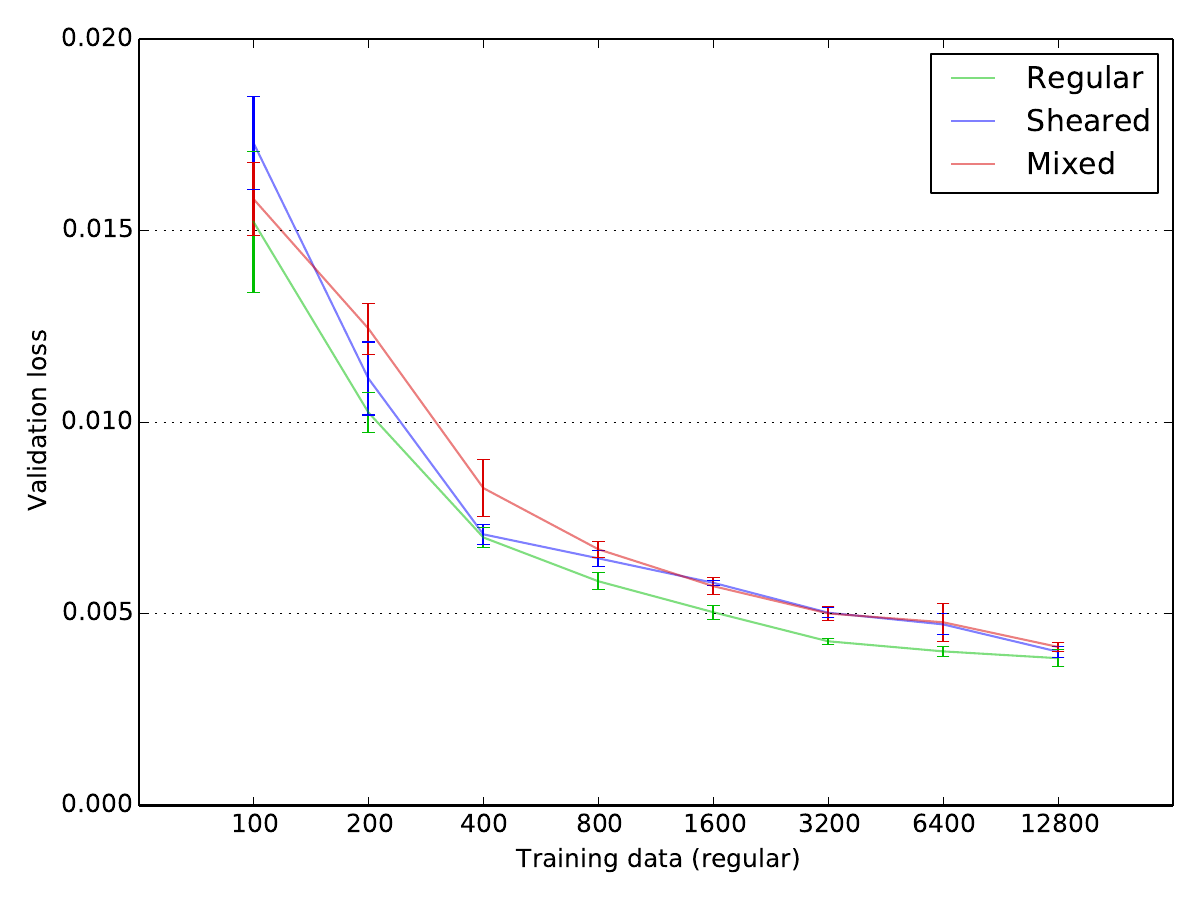}
		\includegraphics[width=0.4\textwidth]{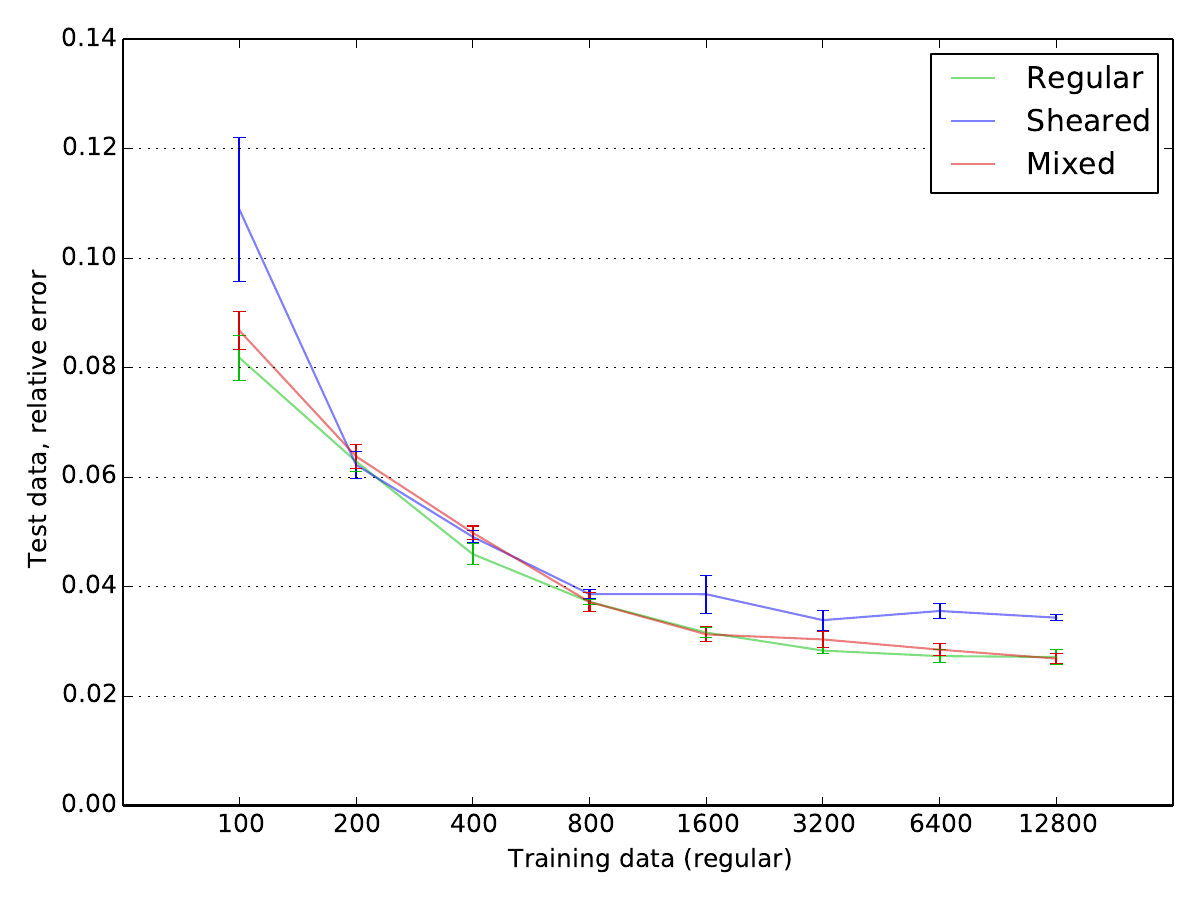}
		\caption{A comparison of validation loss and test errors for different data sets: regular airfoils,
		an augmented set of sheared airfoils, and and a mixed set of data (50\% regular
		and 50\% sheared). All for a model with 7.7m weights. }
		\label{fig:datacomp6}
	\end{center}
\end{figure}

To investigate the generalization behavior 
in more detail, we have 
prepared an augmented data set, where we have sheared the airfoil shapes 
by $\pm 15$ degrees along a centered x-axis to enlarge the space of shapes seen by the networks.
The corresponding relative error graphs for a model with 7.7m weights 
are shown in \myreffig{fig:datacomp6}.
Here we compare three variants, a model trained only with regular data
(this is identical to \myreffig{fig:expcomp}), models purely trained 
with the sheared data set, and a set of models trained with 50\% of 
the regular data, and 50\% of the sheared data. We will refer to these 
data sets as {\em regular}, {\em sheared}, and {\em mixed} in the following.
It is apparent that both the sheared and mixed data sets yield a larger 
validation error for large training data amounts (\myreffig{fig:datacomp6}, top).
This is not surprising, as the introduction of the sheared data leads to an enlarged 
space of solutions, and hence also a more difficult learning task. 
\myreffig{fig:datacomp6} bottom shows that despite this enlarged space, the trained models 
do not perform better on the same test data set from above. 
Rather, the performance decreases when only 
using the sheared data (blue line in \myreffig{fig:datacomp6}, bottom).

\begin{figure}[bt]
	\begin{center}
		\includegraphics[width=0.4\textwidth]{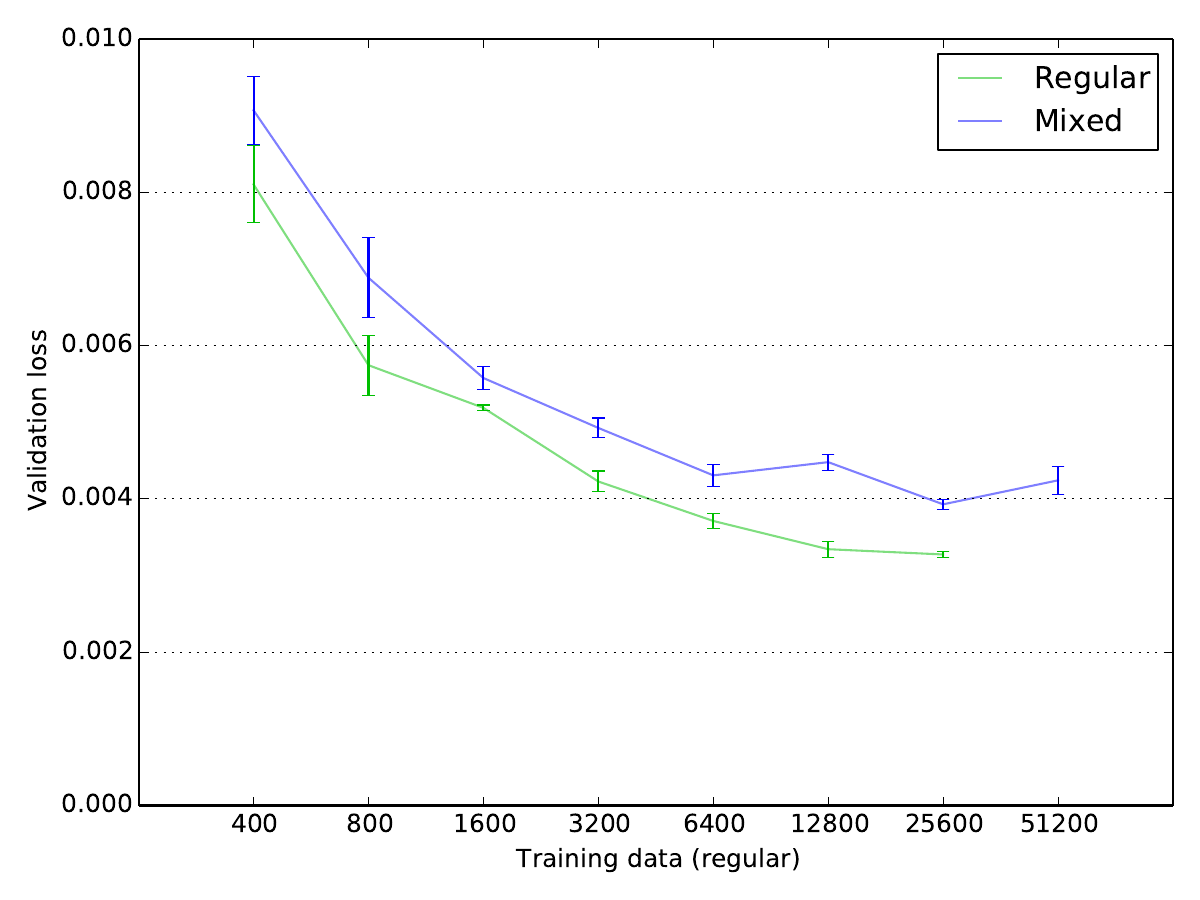}
		\includegraphics[width=0.4\textwidth]{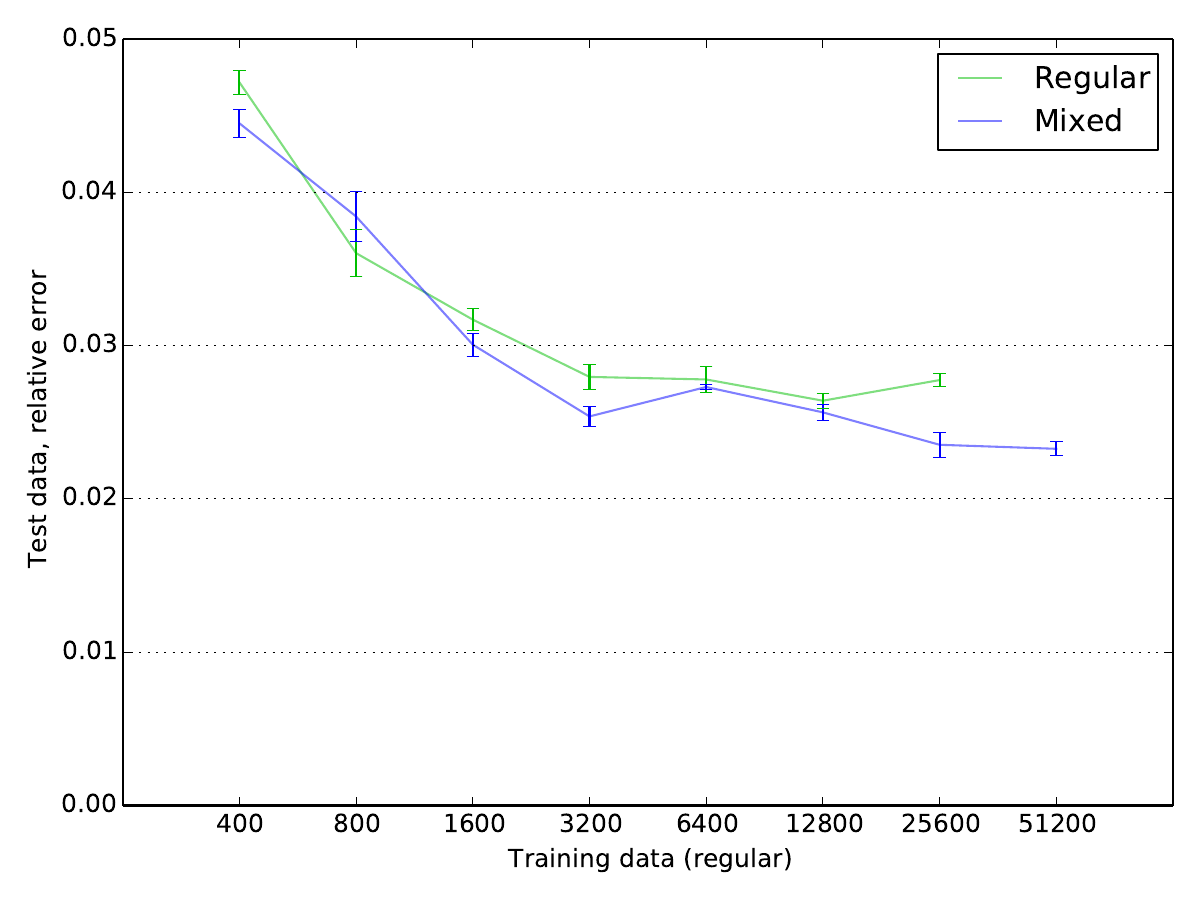}
		\caption{Validation loss and test error for a large model (30.9m weights) trained with regular, 
		and mixed airfoil data. Compared to the smaller model in \myreffig{fig:datacomp6},
		the 30.9m model slightly benefits from the sheared data, 
		especially for the runs with 25.6k and 51.2k data samples.
		}
		\label{fig:datacomp7}
	\end{center}
\end{figure}

\begin{figure*}[tb]
	\begin{center}
	\includegraphics[width=0.99 \textwidth]{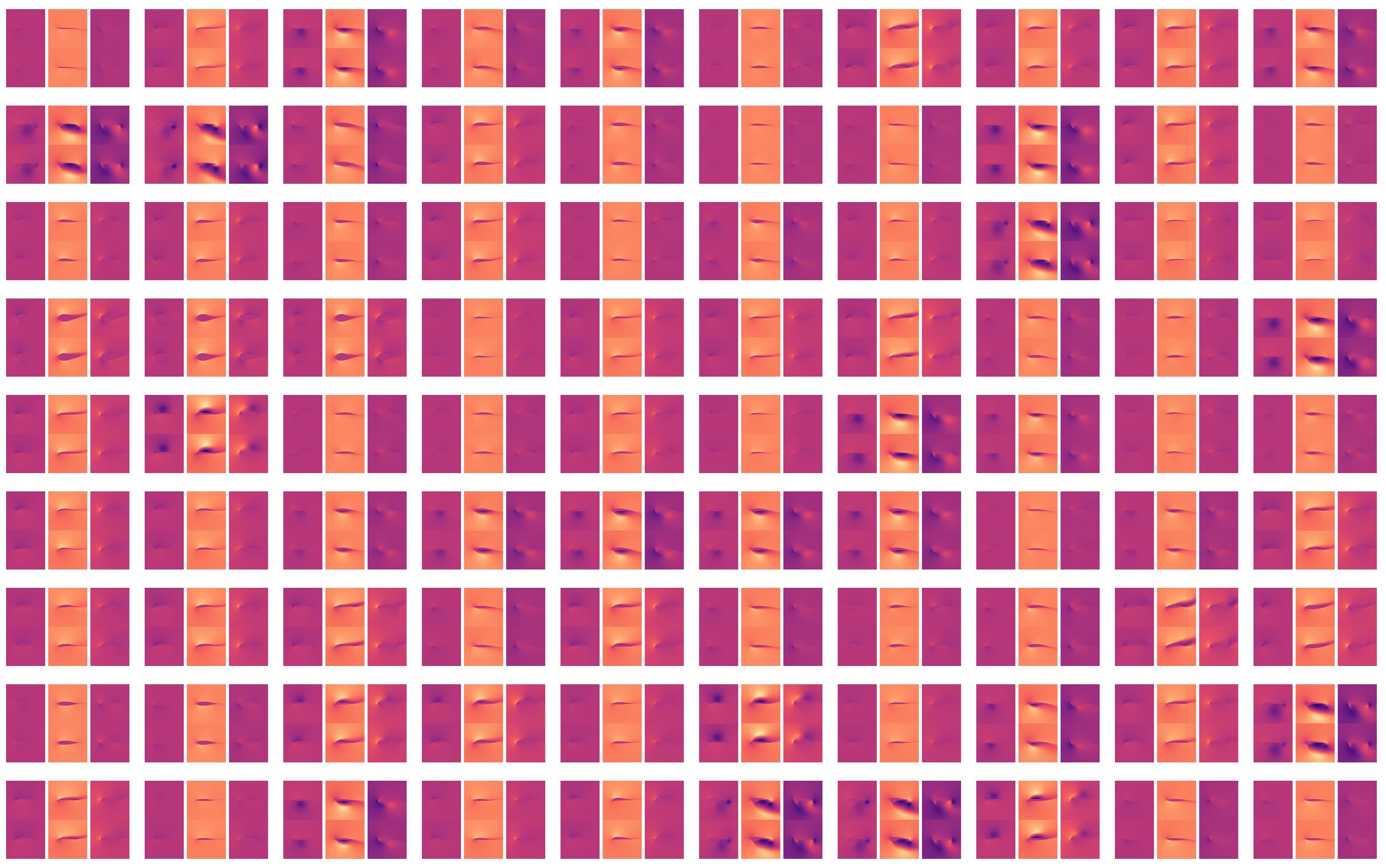}
	\end{center} \vspace{-0.4cm}
	\caption{The full test set outputs from a 30.9m weight model trained with 
	25.600 training data samples. Each image contains reference at the top, and
	inferred solution at the bottom. Each triplet of pressure, x velocity and y velocity
	represents one data point in the test set.
	}
	\label{fig:allResults}
\end{figure*}

This picture changes when using a larger model. Training the 30.9m weight model 
with the mixed data set leads to improvements in performance when 
there is enough data, especially for the run with 25600 training samples in \myreffig{fig:datacomp7}.
In this case, the large model outperforms the regular data model with an average 
relative error of 2.77\%, and yields an error of 2.35\%.
Training the 30.9m model with 51k of mixed samples slightly 
improves the performance to 2.32\%.
Hence, the generalization performance not only depends on type and amount of training data, but also
on the representative capacities of the chosen CNN architecture.
The full set of test data outputs for this model is shown in \myreffig{fig:allResults}.

\paragraph{Performance}

\begin{figure*}[bt]
	\begin{center}
		\includegraphics[width=1.0\textwidth]{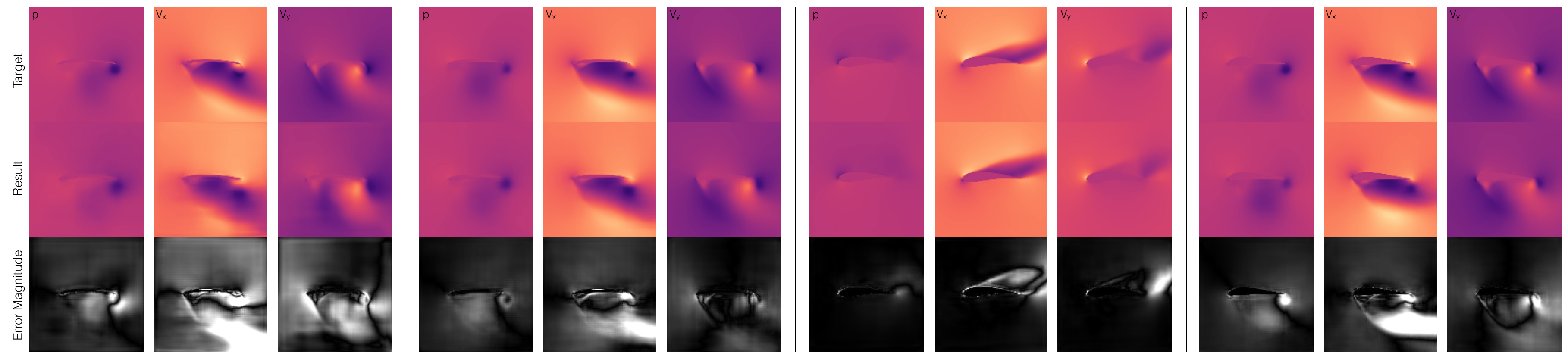}
		\caption{
						A selection of inference test results with particularly high errors.
			Each target triple contains, f.l.t.r., $\hat{p}_o$,$\vti{v}_{o,x}$,$\vti{v}_{o,y}$,
			the model results are shown below. The bottom row shows error magnitudes,
			with white indicating larger deviations from the ground truth targets.
		}
		\label{fig:bad}
	\end{center}
\end{figure*}

The central motivation for deep learning in the context of physics simulations
is arguably performance.
In our case, evaluating the trained 30.9m model for a single data point on an NVidia {\em GTX 1080} GPU 
takes 5.53ms (including data transfer to the GPU). This runtime, like all following 
ones, is averaged over multiple runs. The network evaluation itself, i.e. without 
GPU overhead, takes 2.10ms.
The runtime per solution can be reduced significantly when evaluating multiple solutions at 
once, e.g., for a batch size of 8, the evaluation time rises only slightly to 2.15ms.
In contrast, computing the solution with OpenFOAM requires 40.4s when accuracy is adjusted to match the network outputs. 
The runs for training data generation took 71.9s.

While it is of course problematic to compare implementations as different 
as the two at hand, it still yields a realistic baseline of the performance we 
can expect from publicly available open source solvers.  OpenFOAM
clearly leaves significant room for performance, e.g., its solver is currently
single-threaded. This likewise holds on the deep learning side:
The fast execution time of the discussed architectures can be achieved
''out-of-the-box'' with PyTorch,
and based on future hardware developments such as GPUs with built-in support for NN evaluation, 
we expect this performance to improve significantly even without any 
changes to the trained model itself. 
Thus, based on the current state of OpenFOAM and PyTorch, our models yield a 
speed up factor of ca. 1000$\times$.

The timings of training runs for the models discussed above vary with respect to the amount of 
data and model size, but start with 26 minutes for the 
122k models, up to 147 min. for the 30.9m models.

\paragraph{Discussion}

Overall, our best models yield a very good accuracy of less than 3\% relative error.
However, it is naturally 
an important question how this error can be further reduced. Based on our tests, 
this will require substantially larger training data sets and CNN models.
It also becomes apparent from \myreffig{fig:expcomp} 
that simply increasing model and training data size will not scale to arbitrary
accuracies. Rather, this points towards the need to investigate and develop
different approaches and network architectures.

In addition, it is interesting to investigate how systematic the model errors are. 
\myreffig{fig:bad} shows a selection of inferred results from the 7.7m model trained 
with 12k data sets. All results are taken from the test data set. As can be seen in 
the error magnitude visualizations in the bottom row of \myreffig{fig:bad}, the model 
is not completely off for any of the cases. Rather, errors typically manifest themselves 
as shifts in the inferred shapes of the wakes behind the airfoil. We also noticed that while 
they change w.r.t. details in the solution, the error typically remains large for most of 
the difficult cases of the test data set throughout the different runs. Most likely, this
is caused by a lack of new information the models can extract from the training data sets 
that were used in the study. Here, it would also be interesting to consider an even larger 
test data set to investigate generalization in more detail.

\begin{figure}[bt]
	\begin{center}
		\includegraphics[width=0.5\textwidth]{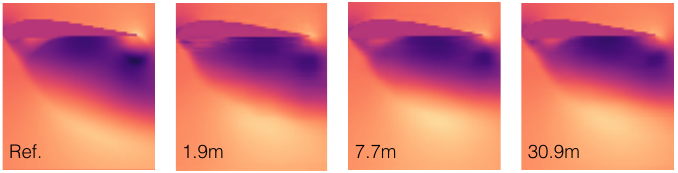}
		\caption{
			A detail of an x velocity, with reference, and three different inferred 
			solutions shown left to right. While the three models have almost identical 
			test error performance, the solutions of the larger networks (on the left)
			are noticeably sharper, e.g., on the left side below the airfoil tip.
			The solutions of the larger models are also smoother in regions further 
			away from the airfoil.
		}
		\label{fig:expcompZoom}
	\end{center}
\end{figure}

Finally, it is worth pointing out that despite the stagnated
error measurements of the larger models in \myreffig{fig:expcomp}, the results 
consistently improve, especially regarding their sharpness.
This is illustrated with a zoom in on a representative case for 
x velocity in \myreffig{fig:expcompZoom}, and we have noticed 
this behavior across the board for all channels and many data sets.
As can be seen there, the sharpness of the inferred function increases,
especially when comparing the 1.9m and 30.9m models, which 
exhibit the same performance in \myreffig{fig:expcomp}. This behavior
can be partially attributed to these improvements being small in terms of scale.
However, as it is noticeable consistently, it also points to inherent limitations 
of direct vector norms as loss functions.

\section{Conclusions}

We have presented a first study of the accuracy of deep learning for
the inference of RANS solutions for airfoils. While the results of our study are by no
means guaranteed to directly carry over to other problems due to the inherent differences 
of solution spaces for different physical problems, we believe that our results can nonetheless 
serve as a good starting point with respect to general methodology, data handling,
and training procedures. 
We hope that they will provide a starting point for researchers,
and help to overcome skepticism from a perceived lack of theoretical results. 
Flow simulations are a good example of a field that has made tremendous steps forward despite 
unanswered questions regarding theory: although it is unknown whether a finite time singularity 
for the Navier-Stokes equations exists, this luckily has not stopped research in the field.
Likewise, we believe there is huge potential for deep learning in the CFD context 
despite the open questions regarding theory.

In addition we have outlined a simulation and training setup that is  
on the one hand relatively simple, but nonetheless offers a large amount of complexity
for machine learning algorithms. 
It also illustrates that a physical understanding of the problem
is crucial, as the nondimensional formulation of the problem leads to significantly improved 
results without any changes to the deep learning components themselves.
I.e., it is important to formulate the problem such that the relationship between 
input and output quantities is as simple as possible.
The proposed setting provides
a good point of entry for CFD researchers to experiment with deep learning algorithms, as well as 
a benchmark case for the evaluation of novel learning methods for fluids and 
related physics problems.

We see numerous avenues for future work in the area of physics-based deep learning, 
e.g., to employ trained flow models in the context of inverse problems. The high performance 
and differentiability of a CNN model yields a very good basis for tough 
problems such as flow control and shape optimization.

\section*{Appendix}

\subsection*{Architecture and Training Details}
\label{app:net}

The network is fully convolutional with 14 layers, 
and consists of a series of convolutional {\em blocks}. 
All blocks have a similar structure:
activation, convolution, batch normalization and dropout.
Instead of transpose convolutions with strides, we use a linear upsampling 
''up()'' followed by a convolution on the upsampled content \cite{odena2016deconv}.
In addition, the kernel size is reduced by one in the decoder part to ensure uneven kernel sizes,
i.e., convolutions with symmetric kernels.

Convolutional blocks $C$ below are parametrized by an output channel factor $c$, kernel size $k$, stride $s$,
where we use $cX$ as short form for $c=X$. Note that the input channels for the decoder part have twice 
the size due to the concatenation of features from the encoder part (this is not explicitly written out in our notation).
Batch normalization is indicated by $b$ below. Activation by ReLU is indicated by $r$, while 
 $l$ indicates a leaky ReLU \cite{maas2013rectifier,RadfordMC15} with a slope of $0.2$. 
Slight dropout with a rate of 0.01 is used for all layers. 
The different models above use a channel base multiplier 
$2^{c_i}$ that is multiplied by $c$ for the individual layers. 
$c_i$ was $3,4,5,6,$ and $7$ for the 
122k, 487k, 1.9m, 7.7m and 30.9m models discussed above. 
Thus, e.g., for $c_i=6$ a $C$ block with $c8$ has $512$ channels.
Channel wise concatenation is denoted by ''conc()''.
{\em Addendum:} Note that $l_5$ below inadvertently used a kernel size of 2 in our original implementation, and is listed
as such here. While for symmetry with the decoder part, $k4$ would be preferable here, 
this should not lead to substantial changes in terms of inference results.

The network receives an input $l_0$ with three channels (as outlined in \myrefsec{sec:prepro}) 
and can be summarized as:

\vspace{-14pt}
\begin{eqnarray}
\small
l_1    &  \leftarrow &            C(l_0,                         c1  \  k4 \  s2          )  \nonumber \\
l_2    &  \leftarrow &            C(l_1,                         c2  \  k4 \  s2 \  l \  b)  \nonumber \\
l_3    &  \leftarrow &            C(l_2,                         c2  \  k4 \  s2 \  l \  b)  \nonumber \\
l_4    &  \leftarrow &            C(l_3,                         c4  \  k4 \  s2 \  l \  b)  \nonumber  \\
l_5    &  \leftarrow &            C(l_4,                         c8  \  k2 \  s2 \  l \  b)  \nonumber  \\
l_6    &  \leftarrow &            C(l_5,                         c8  \  k2 \  s2 \  l \  b)  \nonumber  \\
l_7    &  \leftarrow &            C(l_6,                         c8  \  k2 \  s2 \  l )  \nonumber                 \\
l_8    &  \leftarrow & \text{up}( C( l_7                       , c8  \  k1 \  s1 \  r \  b) )  \nonumber \\
l_9    &  \leftarrow & \text{up}( C( \text{conc}(l_8,l_6)      , c8  \  k1 \  s1 \  r \  b) )  \nonumber \\
l_{10} &  \leftarrow & \text{up}( C( \text{conc}(l_{9},l_{5})  , c8  \  k3 \  s1 \  r \  b) )  \nonumber \\
l_{11} &  \leftarrow & \text{up}( C( \text{conc}(l_{10},l_{4}) , c4  \  k3 \  s1 \  r \  b) )  \nonumber \\
l_{12} &  \leftarrow & \text{up}( C( \text{conc}(l_{11},l_{3}) , c2  \  k3 \  s1 \  r \  b) )  \nonumber \\
l_{13} &  \leftarrow & \text{up}( C( \text{conc}(l_{12},l_{2}) , c2  \  k3 \  s1 \  r \  b) )  \nonumber \\
l_{14} &  \leftarrow & \text{up}( C( \text{conc}(l_{13},l_{1}) ,     \  k3 \  s1 \  r     ) )  \nonumber 
\end{eqnarray}

Here $l_{14}$ represents the output of the network, and the corresponding convolution
generates 3 output channels.
Unless otherwise noted, training runs are performed 
with $i=80000$ iterations 
of the Adam optimizer using $\beta_1=0.5$ and $\beta_2=0.999$,
learning rate $\eta = 0.0004$
with learning rate decay and batch size $b=10$. 
\myreffig{fig:learningRates} used a model with 7.7m weights, $i=40k$ iterations, and 8k training data samples, 75\% regular, and 25\% sheared.
\myreffig{fig:prepro} used a model with 7.7m weights, with 12.8k training data samples, 75\% regular, and 25\% sheared.

\revi{
\subsection*{Training Data}
\label{app:data}
%
In the following, we also give details of the different training data 
set sizes used in the training runs above. We start with a minimal
size of 100 samples, and increase the data set size in factors of two
up to 12800. Typically, the total number of samples is split into 80\% 
training data, and 20\% validation data. However, we found validation sets 
of several hundred samples to yield stable estimates. Hence, we use 
an upper limit of 400 as the maximal size of the validation data set.
The corresponding number of samples are randomly drawn from a  
pool of 26732 pre-computed pairs of boundary conditions and flow solutions computed with OpenFOAM.
The exact sizes used for the training runs of 
\myreffig{fig:learningRates} and \myreffig{fig:expcomp}
are given in \myreftab{tab:data}. The dimensionality of the different 
data sets is summarized in \myreftab{tab:dims}.
\begin{table}[h]
\centering \revi{
\begin{tabular}{|l|l|l|}
\hline
Total dataset size & Training & Validation \\ \hline
100          & 80       & 20         \\ \hline
200          & 160      & 40         \\ \hline
400          & 320      & 80         \\ \hline
800          & 640      & 160        \\ \hline
1600         & 1280     & 320        \\ \hline
3200         & 2800     & 400        \\ \hline
6400         & 6000     & 400        \\ \hline
12800        & 12400    & 400        \\ \hline
25600        & 25200    & 400        \\ \hline
\end{tabular}
\caption{\revi{Different data set sizes used for training runs, together with corresponding
splits into training and validation sets.} \label{tab:data} }
} 
\end{table}
\begin{table}[h]
\centering \revi{
\begin{tabular}{|l|l|}
\hline
Quantity & Dimension \\ \hline
Airfoil shapes (training, validation) &   1505   \\ \hline
Airfoil shapes (test) &   30    \\ \hline
RANS solutions (regular) &   26732    \\ \hline
RANS solutions (sheared) &  27108    \\ \hline
\end{tabular}
\caption{\revi{Dimensionality of airfoil shape database and data sets used for neural network training and evaluation.
} \label{tab:dims} }
} 
\end{table}
}

\revi{
In addition to this regular data set, we employ a {\em sheared} data set 
that contains sheared airfoil profiles (as described above). This data set 
has an overall size of 27108 samples, and was used for the sheared models
in \myreffig{fig:datacomp6} with the same training set sizes as shown in \myreftab{tab:data}.
We also trained models with a {\em mixed} data set,
shown in \myreffig{fig:datacomp6} and \myreffig{fig:datacomp7}
that contains half regular, and half sheared samples. I.e., 
the five models for $x=800$ in \myreffig{fig:datacomp7} employed 
320 samples drawn from the regular plus 320 drawn from the sheared data set for training,
and 80 regular plus 80 sheared samples for validation. For the mixed data set, 
we additionally train a large model with a total dataset size of 25600, i.e., the bottom line of \myreffig{tab:data},
that is shown in \myreffig{fig:datacomp7}.
}

\begin{figure}[bt]
	\begin{center}
		\includegraphics[width=0.3\textwidth]{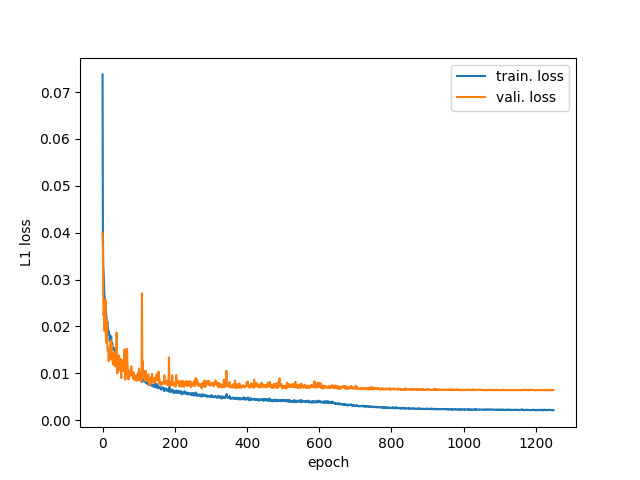}
		\includegraphics[width=0.3\textwidth]{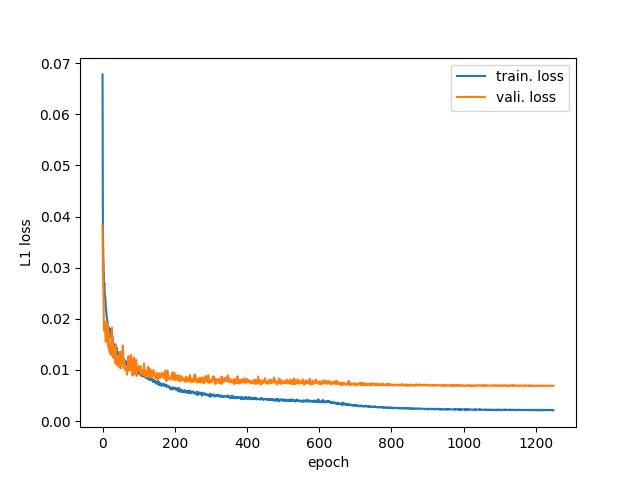}
		\includegraphics[width=0.3\textwidth]{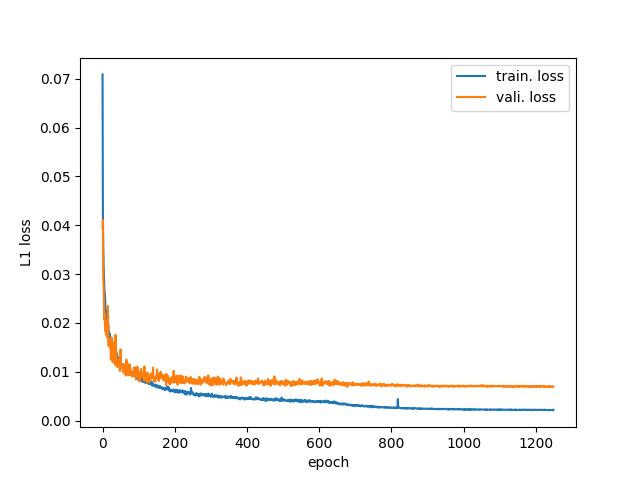}
		\caption{\revi{Training and validation losses for 3 different training 
		runs of a model with 7.7m weights, and 800 training data samples.
		For the x axis, each epoch indicates the processing of 64 samples.}
		\label{fig:losscurves}}
	\end{center}
\end{figure}

\revi{
\subsection*{Training Evolution and Dropout}
\label{app:drop}
In \myreffig{fig:losscurves} we show three examples of training runs of typical model 
used on the accuracy evaluations above. These graphs show that the models converge to stable 
levels of training and validation loss, and do not exhibit overfitting over time.
Additionally, the onset of learning rate decay can be seen in the middle of the graph. This 
noticeably reduces the variance of the learning iterations, and let's the training process 
fine-tune the current state of the model.
}



\begin{figure}[bt]
	\begin{center}
		\includegraphics[width=0.3\textwidth]{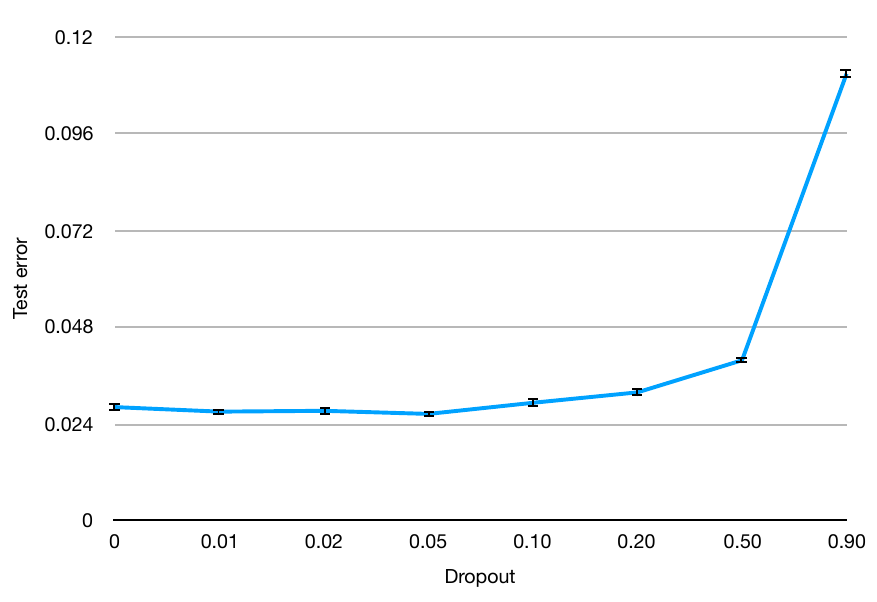}
		\caption{\revi{A comparison of different dropout rates
		for a model with 7.7m weights and 8000 training data samples.}}
		\label{fig:dropout}
	\end{center}
\end{figure}

\myreffig{fig:dropout} evaluates the effect of varying dropout rates.
Different amounts of dropout applied at training time are shown along x 
with the resulting test accuracies. It becomes apparent that the effect is relatively 
small, but overall the accuracy deteriorates with increasing amounts of dropout.

\section*{Funding Sources}

This work is supported by ERC Starting Grant 637014 (realFlow)
and the TUM PREP internship program.

\section*{Acknowledgments}

We thank H. Mehrotra, and N. Mainali for their help with the deep learning experiments,
and the anonymous reviewers for the helpful comments to improve our work.

\small
\bibliographystyle{alpha}
\bibliography{references}

\end{document}